\documentclass[11pt]{article}

\usepackage{acl}
\usepackage{times}
\usepackage{latexsym}
\usepackage[T1]{fontenc}
\usepackage[utf8]{inputenc}
\usepackage{microtype}
\usepackage{booktabs}
\usepackage{multirow}
\usepackage{tabularx}
\usepackage{array}
\usepackage{amsmath}
\usepackage{amssymb}
\usepackage{enumitem}
\usepackage{graphicx}
\usepackage{float}
\usepackage{placeins}
\usepackage{listings}
\usepackage{xcolor}
\usepackage{xspace}
\usepackage{url}
\usepackage{seqsplit}
\newcommand{\shahex}[1]{\begingroup\ttfamily\footnotesize\seqsplit{#1}\endgroup}

\hypersetup{
  pdftitle={BTS-AgentBench: A Deterministic, Replayable Pipeline from Read-Only Telemetry Logs to Agent Benchmarks},
  pdfauthor={Jeong-Yoon Kim and Tae-Yoon Lee}
}

\definecolor{codegray}{gray}{0.96}
\lstdefinestyle{methodpython}{
  language=Python,
  basicstyle=\ttfamily\footnotesize,
  breaklines=true,
  columns=fullflexible,
  keepspaces=true,
  frame=single,
  backgroundcolor=\color{codegray},
  showstringspaces=false
}
\lstdefinestyle{traceblock}{
  basicstyle=\ttfamily\footnotesize,
  breaklines=true,
  columns=fullflexible,
  keepspaces=true,
  frame=single,
  backgroundcolor=\color{codegray},
  showstringspaces=false
}

\title{BTS-AgentBench: A Deterministic, Replayable Pipeline from Read-Only Telemetry Logs to Agent Benchmarks}

\author{Jeong-Yoon Kim \quad Tae-Yoon Lee \\
Department of Electronic Engineering \\
Hanbat National University}

\begin{document}
\maketitle

\begin{abstract}
Operational facilities accumulate extensive local telemetry, but agent-ready interactive evaluations with source-verifiable answers are costly to construct from such archives. Site-specific measurement histories also motivate a local conversion workflow that can operate where raw operational data must remain under organizational control. We introduce BTS-AgentBench, a deterministic compiler that transforms read-only telemetry and equipment metadata into tool-backed, multi-turn agent tasks, phase-level targets, evidence and executable evaluation contracts. The BTS release includes 532 episodes across nine families; a second-corpus port produces 204 XAI4HEAT episodes. Frozen-input replay reproduces released artifacts byte-for-byte, and independent raw-source checks cover all 89 BTS and 41 XAI4HEAT test episodes. Replayable provenance links 1,170 trial identities to recorded execution and scorer conditions, while 1,125 completed traces retain model responses and tool results. An obligation-level comparison of 216 scored failures found that linked records exposed additional recorded failed obligations in 202 cases (93.5\%), including tool arguments, stream identity, state carryover and missing interaction steps. The pipeline thereby supports source-grounded task generation, exact reconstruction and auditable diagnosis of agent executions.
Code, released benchmarks, and reproduction instructions are available at \href{https://anonymous.4open.science/r/telemetry-benchmark-reproduction-3D68/}{\nolinkurl{https://anonymous.4open.science/r/}\allowbreak\texttt{\seqsplit{telemetry-benchmark-reproduction-3D68/}}}.
\end{abstract}

\section{Introduction}

Modern buildings and industrial facilities retain substantial histories of measurements, equipment metadata and operating conditions. Public datasets such as BTS and XAI4HEAT already make these observations available for research \citep{prabowo2024bts,cvetkovic2025xai4heat}. Yet raw time series alone do not specify what an agent should ask, which streams its tools should retrieve, how a revised request changes the calculation, or what evidence makes an answer verifiable. Constructing such tasks manually for each site and sensor family requires repeated specification of identities, temporal semantics, dialogue transitions and answer criteria. A local data-to-benchmark compiler can process site-specific records under organizational data-access controls.

We address this gap with a deterministic procedure that converts telemetry and equipment metadata into tool-backed tasks, multi-turn interactions, source-derived reference calculations and executable evaluation contracts. The compiler couples the operator-facing request, simulator state, tool behavior, intermediate answers and scoring obligations to the same observation history. An agent can thus be evaluated on clarification, goal revision, timestamp and quality decisions, and evidence follow-ups rather than on a one-shot numerical answer alone.

A final numerical answer does not necessarily reveal whether an agent queried the intended sensor, carried revised context forward, or completed the required evidence follow-up. These obligations are consequential for tool-using analysis: a plausible report can be grounded in the wrong stream or omit an earlier decision. BTS-AgentBench therefore preserves the source observations, interaction contracts, chronological tool calls, phase-level checks and evaluation records together.

This linkage makes intermediate failure obligations observable. Among 216 scored failures in the paid-run archive, 123 (56.9\%) passed the final-answer component. Final-answer checks alone identified all recorded failed obligations in 14 cases (6.5\%); linked records exposed additional failures in 202 (93.5\%). The linked records identified phase failures in 181 cases and execution-check failures in 132. With 1,170 trial identities and 1,125 completed execution traces linked to source and evaluator conditions, this evidence can be inspected and rescored from preserved execution artifacts.

BTS supplies public multi-year time series and standardized metadata \citep{prabowo2024bts}. A Telemetry Adapter normalizes source identities and observations; a Corpus Contract declares tool, temporal and quality semantics. The downstream compiler adds clarification, revision, commitment and evidence while preserving source computation and split. XAI4HEAT tests reuse across row-oriented SCADA rather than graph-plus-per-stream archives, retaining shared components with two localized interaction/reportability adaptations.

The telemetry-to-episode construction method is the primary contribution; BTS-AgentBench is its released benchmark instance. Our contributions are:
\begin{itemize}[leftmargin=1.25em]
    \item A typed compilation interface that separates normalized telemetry ingestion from reusable task, interaction, replay, and evaluator components.
    \item A deterministic compiler that transforms normalized telemetry into executable task candidates, typed multi-turn contracts, tool-derived intermediate/final golds, evidence, and verifiers.
    \item Three distinct replay guarantees---selection replay, artifact replay, and trajectory rescoring---with an end-to-end BTS release and explicit replay boundaries.
    \item A secondary XAI4HEAT port regenerating 204 episodes with shared protocol/runtime/scorer/replay components and corpus-specific interaction/reportability adaptation; an independent audit checks all 41 test rows without mismatch, demonstrating cross-corpus reuse.
    \item A provenance-preserving evaluation workflow that links archived agent actions to source-derived targets and identifies failed intermediate obligations.
\end{itemize}

We evaluate construction, source grounding, frozen-input replay, cross-corpus reuse, and archived-trace scoring. The episodes cover clarification, goal revision, temporal and quality policies, and evidence-backed reporting.

\section{Related Work}

\paragraph{Interactive tool-use benchmarks.}
BTS-AgentBench is closest to benchmarks separating an agent, user surface, tools, and programmatic evaluation. API-Bank provides runnable tool-use dialogues \citep{li2023apibank}; ToolSandbox adds state and implicit dependencies \citep{lu2025toolsandbox}; $\tau$-bench combines simulated users, APIs, policies, and goal scoring \citep{yao2025taubench}; and ACEBench adds granular multi-turn error dimensions \citep{chen2025acebench}. GTA studies tool use on real-world user queries and deployed tools \citep{wang2024gta}. BTS-AgentBench compiles site-specific telemetry into coordinated executable tasks, interaction targets, and source-derived reference answers while preserving their reconstruction chain.

\paragraph{Operational and industrial agent benchmarks.}
ITBench \citep{jha2025itbench} and AssetOpsBench \citep{patel2026assetopsbench} evaluate structured operational workflows; ReAct Meets Industrial IoT \citep{rayfield2025react} studies language-agent access to industrial telemetry. BTS-AgentBench complements them by focusing on compilation of read-only telemetry into bounded, deterministically evaluated episodes.

\paragraph{Building telemetry data and metadata.}
BTS provides real building time series with standardized metadata \citep{prabowo2024bts}, while Brick represents building sensors, equipment, and relationships portably \citep{balaji2018brick}. We use this substrate to derive executable telemetry-analysis tasks and agent episodes.

\begin{figure*}[t]
    \centering
    \small
    \begin{tabularx}{0.88\textwidth}{>{\centering\arraybackslash}X}
    \fbox{\begin{minipage}{0.82\textwidth}\centering
    \textbf{Raw building telemetry}\quad Equipment/point metadata and timestamped raw values; e.g., BTS\_C Zone 005, $12.9457$ at 2022-02-03 07:03:23.640 UTC.
    \end{minipage}} \\
    $\Downarrow$ \\
    \fbox{\begin{minipage}{0.82\textwidth}\centering
    \textbf{Telemetry Adapter}\quad Corpus-specific parsing and mapping to normalized site, stream, point, equipment/location, time, value, and source-lineage fields.
    \end{minipage}} \\
    $\Downarrow$ \\
    \fbox{\begin{minipage}{0.82\textwidth}\centering
    \textbf{Corpus Contract}\quad Declared identity, UTC/window, missing-value, tool, aggregation, nearest, quality, and tie semantics.
    \end{minipage}} \\
    $\Downarrow$ \\
    \fbox{\begin{minipage}{0.82\textwidth}\centering
    \textbf{Deterministic episode compiler}\quad Select source-grounded task; compose dialogue and state revisions; execute tools to derive phase golds and evidence.
    \end{minipage}} \\
    $\Downarrow$ \\
    \fbox{\begin{minipage}{0.82\textwidth}\centering
    \textbf{Replayable interaction artifact}\quad User turns, tool contract, source-derived intermediate results, state transitions, evidence, and final-answer targets.
    \end{minipage}} \\
    $\Downarrow$ \\
    \fbox{\begin{minipage}{0.82\textwidth}\centering
    \textbf{Evaluator}\quad Offline checks for task outputs, interaction phases, temporal/state behavior, evidence, and protocol.
    \end{minipage}} \\
    $\Downarrow$ \\
    \fbox{\begin{minipage}{0.82\textwidth}\centering
    \textbf{Optional controller/model execution}\quad Produces the execution trace consumed by the evaluator.
    \end{minipage}}
    \end{tabularx}
    \caption{Telemetry-to-benchmark pipeline. The concrete source value follows the final-artifact worked example in Table~\ref{tab:episode-compilation}; numeric values are raw and unitless unless the source supplies authoritative units.}
    \label{fig:overview}
\end{figure*}

\paragraph{Time-series and executable benchmark construction.}
TimeSage-MT constructs verifiable multi-turn reasoning tasks from real-world time series across application domains \citep{kong2026timesagemt}. BTS-AgentBench focuses on local telemetry identity, read-only tools, typed temporal and quality policies, and provenance-preserving artifact and scorer replay. SUPER compiles executable research-repository tasks \citep{bogin2024super}, providing a complementary instance of evaluations grounded in executable artifacts.

\section{Benchmark Formulation}

Figure~\ref{fig:overview} gives an overview of BTS-AgentBench, from building telemetry to released multi-turn agent episodes. The figure also illustrates the evaluation interface used throughout the paper: an operator-facing request, read-only tool use, a policy decision for nearest observation, and an evidence-backed answer.

\subsection{Benchmark Unit}

BTS-AgentBench first compiles telemetry and metadata into executable read-only tasks, then converts each retained task into a bounded episode. A source task $r$ records its family, split, tools, evidence, and gold contract; $r^\star$ preserves its computation and target while changing the agent-facing interface.

Each episode combines a deterministic user simulator, read-only tools, an agent trace, and a programmatic evaluator. The deterministic evaluator checks five aspects of each execution: the required task result and grounded tool behavior, intermediate and revised-state obligations, source evidence, the terminal answer or commitment, and completion of the required interaction graph. A row is counted as complete only when every applicable mandatory check passes. Exact evidence-to-result provenance is tracked separately because ordinary execution traces identify the cited stream but do not uniquely identify the originating call/result for every evidence item.

\paragraph{Construction, scoring, and audit roles.}
The construction pipeline and evaluator occupy distinct stages of the benchmark lifecycle (Table~\ref{tab:role-separation}). The \textbf{compiler} creates benchmark rows and source-derived golds from frozen telemetry. The \textbf{evaluator} scores produced traces against those executable contracts. 

\begin{table}[!htbp]
\centering
\footnotesize
\setlength{\tabcolsep}{3pt}
\renewcommand{\arraystretch}{1.06}
\begin{tabularx}{\columnwidth}{>{\raggedright\arraybackslash}p{0.20\columnwidth}>{\raggedright\arraybackslash}p{0.31\columnwidth}>{\raggedright\arraybackslash}X}
\toprule
Component & Input / output & Role in the benchmark \\
\midrule
Compiler & Raw telemetry + metadata $\rightarrow$ episode contract, tool-derived golds, evidence & Builds the benchmark before evaluation. \\
Evaluator & Agent trace + frozen episode contract $\rightarrow$ scores & Checks observable behavior and task completion offline. \\
Controller & Revealed dialogue + public tools + its own state/results $\rightarrow$ trace & Produces deterministic audit traces for programmatic-compressibility analysis. \\
\bottomrule
\end{tabularx}
\caption{Separation between benchmark construction and scoring.}
\label{tab:role-separation}
\end{table}

\subsection{Why Replayability?}
\label{sec:why-replay}
Replayability is the mechanism that links raw observations to retained benchmark targets and scores. We distinguish three operations because they answer different audit questions.

\textbf{Selection replay} starts from the frozen corpus inputs, normalized adapter output, Corpus Contract, and compiler configuration, then reruns candidate enumeration, eligibility filters, split assignment, and episode selection to recover the same selected identities and order. For BTS, the replay boundary begins at the frozen normalized catalog, which fixes graph-to-catalog representative choices.

\textbf{Artifact replay} starts from selected episode identities and recomputes point resolution, time windows, tool outputs, intermediate states, phase/final golds, evidence, and verifier targets from source telemetry. It verifies the source-to-gold path for every retained row.

\textbf{Trajectory rescoring} applies the frozen evaluator to retained messages, calls, and outputs without rerunning inference. It reproduces the score associated with a stored trace and contract.

Together, the three replay modes provide a reproducible path from selected source observations to benchmark artifacts and from stored executions to scores, while localizing discrepancies to selection, construction, or evaluation.

\subsection{Telemetry Adapter and Corpus Contract}
\label{sec:adapter-contract}
The compiler consumes a normalized telemetry interface independent of archive layout. A \textbf{Telemetry Adapter} reads corpus records and metadata and emits normalized point identities and observations with site key, stable stream key, point class, equipment/location descriptors, timestamp, raw numeric value, and source lineage. Engineering units are carried through when supplied by authoritative source metadata.

A \textbf{Corpus Contract} is the versioned declarative specification consumed by the compiler and tool runtime. It defines schema and identity rules, timezone and interval conventions, missing-value and aggregation behavior, read-only operations and their argument/result schemas, and deterministic policies for exact/nearest lookup, quality decisions, comparisons, and ties. It also selects the task families supported by the corpus. Adapter and contract together isolate source interpretation from the reusable construction and evaluation stack.

For BTS, the adapter matches graph-derived stream metadata to raw archive histories. For XAI4HEAT, a separate adapter parses row-oriented SCADA tables and maps substation--channel pairs into the same point/stream/time/value representation. Porting XAI4HEAT required localized adaptation of interaction grammar, revision logic, and reportability/quality construction in two corpus-specific adaptation modules. The normalized interface, candidate builders, dialogue renderer, clarification logic, tool runtime, protocol dependency construction, deterministic scorer, trace alignment, and replay logic were reused unchanged or through configuration. The final BTS artifact reports source numeric values directly; engineering units are attached only when the source metadata supplies them.

\subsection{Static Executable Task Layer}
\label{sec:static-layer}

BTS itself is a building time-series corpus rather than a pre-existing suite of the telemetry-analysis tasks evaluated here. Its Telemetry Adapter joins processed BTS metadata to raw archive histories, normalizes site, point, equipment, and location fields, records candidate ambiguity, and scans matched streams for timestamp coverage. Only metadata points matched to raw history become \emph{tool-ready points}. Under the BTS Corpus Contract, these normalized records are materialized in a DuckDB-backed read-only tool store with a raw stream index, per-stream quality statistics, daily/weekly/monthly aggregates, calendar profiles, and previews. Runtime operations use the units and semantics explicitly available in the normalized source metadata.

Family builders apply fixed eligibility predicates before natural-language rendering. Point-disambiguation cases require at least two same-site streams of one point class while the target metadata tuple resolves uniquely. Aggregate candidates below the corpus 10th-percentile coverage floor or above the site/class 99.5th-percentile absolute-mean cap are removed. Pairwise candidates compare distinct equipment in one site/class/window and require a positive difference at or above the corresponding median; rank candidates require at least three streams and a positive top-to-runner-up margin at or above the group median. Exact-timestamp tasks select an interior observation, nearest-timestamp tasks place the request inside a typical adjacent-observation interval, and quality tasks execute fixed weekly coverage and gap rules derived from corpus quantiles.

\begin{table}[!htbp]
\centering
\footnotesize
\setlength{\tabcolsep}{4pt}
\renewcommand{\arraystretch}{1.08}
\begin{tabularx}{\columnwidth}{>{\raggedright\arraybackslash}X
                                >{\raggedleft\arraybackslash}p{0.24\columnwidth}
                                >{\raggedleft\arraybackslash}p{0.18\columnwidth}}
\toprule
Source category & Candidate space & Retained tasks \\
\midrule
BTS metadata streams & 19,665 & -- \\
Tool-ready points & 14,422 & -- \\
\midrule
Point lookup/disambiguation & 4,263 & 60 \\
Day mean lookup & 2,193,431 & 60 \\
Relative 24h mean lookup & 2,193,431 & 60 \\
Window mean lookup & 315,929 & 53 \\
Window pairwise compare & 5,989,083 & 60 \\
Window rank & 1,084 & 60 \\
Timestamp value lookup & 2,123 & 60 \\
Timestamp nearest lookup & 2,123 & 60 \\
Quality-aware reporting & 315 & 59 \\
\midrule
Total retained rows & -- & 532 \\
\bottomrule
\end{tabularx}
\caption{Source pool and retained static task set derived from BTS data before episode lifting. Candidate counts are induced by the tool-store construction and task generator; the retained tasks use a 356/87/89 train/dev/test split.}
\label{tab:funnel}
\end{table}

Candidate counts in Table~\ref{tab:funnel} are combinatorial and often share streams, equipment groups, or nearby correlated windows. We therefore reserve BTS\_C for testing and retain a diversity-capped panel across point class and calendar quarter, adding location type for ranking and decision label for quality-gate tasks. Per-family targets are 40/10/10 train/dev/test, except quality gate (40/10/9) and window mean (36/7/10), where eligible balanced candidates are exhausted. Every fifth deterministically ordered non-held-out candidate is assigned to development and the remainder to training. These rules were fixed before model evaluation; model output and the descriptive difficulty proxy play no role in retention or split assignment. The family generators require raw-backed streams, resolvable point identities, valid time windows, finite tool-derived reference values, and complete tool-call, gold, evidence and verifier fields. Point disambiguation uses same-site class ambiguity with a unique full descriptor; means require available aggregates; pairwise and rank tasks exclude ties and apply margin rules; timestamp tasks use source observations or deterministic between-observation requests; quality tasks require computable coverage and gap ratios. Within each split, diversity caps balance point class, calendar quarter, and, where relevant, location type and answer/abstain labels. The frozen selection contract matches all 532 selected identities to regenerated source-derived task signatures.

Each retained row records its query, canonical calls, acceptable alternatives, structured gold, stream evidence, verifier, and source metadata. Alternatives are generated by builder rules, including location omission only when uniqueness is preserved, equivalent period encodings, consistently reversed pairwise arguments, and exact-to-nearest fallback where allowed. Pairwise and rank ties are excluded by positive margins; stream ID is used only as a final deterministic ordering key, and nearest lookup selects the earlier observation when absolute offsets are equal. Gold fields are obtained from the same read-only store or referenced raw history and are independently checked before and after episode construction. A frozen 532-identity selection contract links regenerated candidates to the evaluated release without copying any conversation or model response.

\section{From Static Tasks to Agent Episodes}

\subsection{Deterministic Episode Compilation}

The compiler fills typed task fields from observations and metadata, not blanks in a prewritten assistant conversation. Metadata binds site, equipment, point class and stream; a family rule selects an operation and its time arguments; read-only tools derive the numeric gold and evidence. Interaction rules then withhold a designated user-context field, generate its clarification reply, and specify later revisions and quality/evidence requests. For example, the source time 07:03:23.640 is stored privately while the initial request omits time. Only a matching clarification exposes it; a later request changes the time policy while retaining the stream. Family rules and templates are authored once, not per row. Released episodes store concrete user turns, reveal/transition rules and phase-bound golds, not generated model answers. During evaluation the agent chooses its own questions, calls and answers; the simulator advances conditionally. Construction is deterministic, but stochastic model outputs are not regenerated by artifact replay.

Figure~\ref{fig:raw-to-episode} traces one released row; Appendix~\ref{sec:final-compilation-detail}, Table~\ref{tab:episode-compilation}, gives its field-level construction. The revised public minute comes from the source timestamp, the quality window is its calendar week, and evidence retains the static task's stream. Appendix~\ref{sec:exact-row-replay} gives the historical conversation and replay example.

\begin{figure*}[t]
\centering
\scriptsize
\setlength{\tabcolsep}{6pt}
\renewcommand{\arraystretch}{1.1}
\begin{tabularx}{\textwidth}{r>{\raggedright\arraybackslash}X}
1 & \textbf{Raw building telemetry:} \,\texttt{Site\_Caa/2254.pickle} contains stream \texttt{c24589e8...9d20}; at 2022-02-03 07:03:23.640 UTC its raw value is 12.9457. \\
$\downarrow$ & \\
2 & \textbf{Point/equipment catalog:} the stream maps to BTS\_C, Zone 005, air differential pressure. \\
$\downarrow$ & \\
3 & \textbf{Source-grounded task candidate:} resolve that descriptor, then request the exact observation at the recorded timestamp. \\
$\downarrow$ & \\
4 & \textbf{Episode contract:} hide the time initially; retain the resolved stream and bind exact-value and evidence targets. \\
$\downarrow$ & \\
5 & \textbf{Multi-turn dialogue:} ask for the missing time; after clarification, revise the request to the nearest reading around 07:03. \\
$\downarrow$ & \\
6 & \textbf{Tool runtime:} resolve point; exact lookup; nearest lookup; inspect quality for the week beginning January 31. \\
$\downarrow$ & \\
7 & \textbf{Gold intermediate results:} exact and nearest value 12.9457; nearest offset 23.64 seconds; weekly coverage 1.0 and gap ratio 1.0563. \\
$\downarrow$ & \\
8 & \textbf{Evidence and decision contracts:} source stream identity; quality decision \texttt{answer}; reporting reason \texttt{nearest\_but\_acceptable}. \\
$\downarrow$ & \\
9 & \textbf{Replayable episode:} regenerate revealed turns, executed tool outputs, state, and gold targets from frozen source data and rules. \\
$\downarrow$ & \\
10 & \textbf{Offline evaluator:} score tool grounding, revised-state use, result-conditioned commitment, evidence, and protocol separately. \\
\end{tabularx}
\caption{End-to-end construction of \texttt{test\_timestamp\_value\_lookup\_00051}. The dialogue revision and later reportability decision are derived from the source task and deterministic tool results; evaluated agents produce their own calls and responses.}
\label{fig:raw-to-episode}
\end{figure*}

\paragraph{Surface realization.} Mode assignment is a typed branch over the task family and its primary telemetry call. Point-disambiguation and quality rows expose site as the missing slot; exact-timestamp and aggregate rows expose time; nearest-fallback rows retain a public timestamp but require nearest-observation handling; the remaining rows use a direct request. The compiler removes the designated site or temporal span from the static query, selects a deterministic family-aware operator wrapper, and fills clarification responses from typed fields. Dates, trailing-24-hour anchors, week or month starts, and fractional-second timestamps each have explicit renderers. References such as ``the same signal,'' ``the next day,'' or ``the second month's winner'' are emitted from stored stream and phase bindings, so their antecedents and revised operations remain explicit in the contract. Mode selection, surface templates, and missing values are therefore generated entirely from the typed contract.

\paragraph{Obligation composition.} Added turns are constructed from tool-valid structured state rather than appended independently as text. For the row in Table~\ref{tab:episode-compilation}, the timestamp rule floors the exact source time to a minute and probes candidate minutes in the fixed order $0,-1,+1,-2,+2,-3,+3$. It selects the first candidate with no exact observation and a defined nearest observation, then stores the requested and observed timestamps, value, fallback reason, and offset while attaching the exact and nearest calls. The quality rule executes \texttt{inspect\_quality\_window} over the declared calendar window and stores its decision, coverage, and gap statistics. A fixed reporting rule then derives \texttt{answer}, \texttt{abstain}, or \texttt{re\_clarify} from the latest timestamp and quality phases. Each obligation-bearing transform updates the user turn, milestone, canonical calls, phase gold, final target, evidence, and verifier together.

\paragraph{Simulator execution.} At evaluation time, the user simulator is a finite-state process with pending clarification slots, an initial-answer flag, a revision index, and a post-answer index. Tool-call messages do not advance user state. A site or time value is released only after the agent asks a matching clarification question; otherwise the simulator requests the still-missing detail. After the first substantive answer, it emits the stored goal revisions in order, followed by any rationale and evidence requests, and then terminates. If the agent asks another question during a revision or evidence phase, the simulator restates the fixed context instead of inventing new information. Consequently, the observed conversation depends on the agent's behavior, while the user information state and executable targets remain identical across systems.

\begin{table}[!htbp]
\centering
\footnotesize
\setlength{\tabcolsep}{4pt}
\renewcommand{\arraystretch}{1.08}
\begin{tabularx}{\columnwidth}{>{\raggedright\arraybackslash}p{0.36\columnwidth}>{\raggedright\arraybackslash}X}
\toprule
Phase type & Evaluated role \\
\midrule
Clarification & Recover missing site or time context before querying. \\
Initial answer & Execute the source telemetry task with read-only tools. \\
Goal revision & Reuse resolved state under a revised request. \\
Timestamp policy & Decide exact, nearest, or insufficient-time reporting. \\
Quality commitment & Answer or abstain under quality evidence. \\
Rationale follow-up & Explain a quality or reportability decision. \\
Evidence follow-up & Return the stream, point, timestamp, or aggregate evidence supporting the answer. \\
\bottomrule
\end{tabularx}
\caption{Canonical phase vocabulary used to evaluate converted agent episodes.}
\label{tab:phase-schema}
\end{table}

\subsection{Canonical Episode Schema}

Each canonical phase contains a user turn, required state update, allowed tool-grounded behavior, and a typed contract target. Families instantiate only phases required by their telemetry operation. Mandatory obligations are checked from ordinary execution events where observable; the remaining evidence-to-result provenance boundary is reported separately.

Point and timestamp families combine clarification, timestamp-policy, and evidence phases; aggregate, comparison, and ranking families add revised-window computation; and quality-gate rows add commitment and rationale phases. Each released row records the applicable ordered subset rather than forcing every family through an identical dialogue.

\subsection{Typed Contract Composition and Repair}

\paragraph{Contract representation.}
Each operator turn is paired with an executable obligation. The compiler represents an episode as
\begin{equation}
C=(Q,\Phi,A,E,V), \qquad \phi_i=(f_i,g_i,R_i),
\end{equation}
where $Q$ stores the interaction mode, missing slots, and ordered operator turns; $\Phi=(\phi_1,\ldots,\phi_n)$ is the phase sequence; $f_i$ is a phase type from Table~\ref{tab:phase-schema}; $g_i$ is its structured gold map; and $R_i$ lists the fields required for scoring. $A$ contains canonical and acceptable read-only tool calls, $E$ identifies the contributing streams, and $V$ contains milestones, tolerances, and protocol conditions. For example, a timestamp phase requires an observed timestamp and value, whereas a quality phase requires a decision, coverage, and gap statistic. The static task supplies the first executable phase. Authored templates render $Q$ and $\Phi$ into operator-facing turns, while $\Phi$, $A$, $E$, and $V$ retain the machine-checkable contract behind those turns. This separation permits programmatic generation at telemetry scale: natural-language fields are instantiated from a finite typed contract, while values, times, streams, and decisions remain executable data.

Protocol dependencies are generated from semantic slot consumption: a clarification must precede the first operation that consumes the clarified slot, while independent operations may occur in either order. At evaluation time, ordinary dialogue, tool calls, and results are deterministically aligned to the typed contract; agents do not emit evaluator-internal identifiers.

\paragraph{Deterministic contract transformation.}
The compiler applies an ordered sequence of typed transformations whose predicates are evaluated over structured contract fields. Applied stages execute declared read-only operations and update the affected turn, phase state, tool calls, golds, evidence, and verifier together. Appendix~\ref{sec:contract-transformation-details} gives the formal transformation rule and coupled preservation/grounding conditions.

Detailed executable transformation and audit-trail examples are provided in Appendix~\ref{sec:contract-transformation-details}.

\section{Evaluation Setup}

\subsection{Final Artifact Validation}
\label{sec:artifact-validation}
The empirical object is the final benchmark artifact. It contains 532 rows across nine families with a 356/87/89 train/dev/test split. Contract preflight reports zero coded issues, and all 89 canonical test traces pass the final deterministic scorer.

Selection and artifact replay start from checksummed BTS archives, a frozen normalized catalog, the revision configuration, and a 532-identity selection manifest. The replay rebuilds the read-only tool store, regenerates selected static tasks and episodes, and byte-compares all split files; the recorded clean replay is byte-identical. The normalized catalog defines the replay boundary because graph metadata can map multiply-related streams to several possible point/equipment representatives. Appendix~\ref{sec:exact-row-replay} documents the commands and hashes.

The review artifact provides raw-download and reconstruction commands, the released splits including semantic provenance, and offline scoring replay.\footnote{\url{https://anonymous.4open.science/r/telemetry-benchmark-reproduction-3D68/}}

\section{Results}
\subsection{RQ1: Candidate Space and Released Evaluation Panel}
BTS metadata contains 19,665 streams, of which 14,422 match raw archive histories and enter the tool-ready inventory. Nine family builders enumerate source-executable candidates and apply identity, data availability, unambiguous gold, coverage and quality rules. Table~\ref{tab:funnel} reports the resulting candidate and retained task counts. The release selects 532 tasks under fixed family quotas, site-held-out splitting, and within-split diversity caps. The BTS\_C test site spans 17 point classes, 26 equipment labels, 37 evidence streams, and source times from May 2020 to January 2024.

\subsection{RQ2: Interaction Structure Added by Compilation}
The 89-row test panel has 3--5 gold phases per row (15, 22, and 52 rows, respectively; mean 4.42). Its compiled contracts contain 79 rows with clarification obligations, 89 with at least one goal revision, 304 total revision turns, 98 post-answer follow-up turns, and 451 canonical tool calls (mean 5.07 per row). Every row includes a source-evidence target; quality is checked in every row, and 79 rows require missing-context recovery.

Family-level interaction counts appear in Appendix~\ref{sec:interaction-structure-details}.

The compilation links each interaction requirement to a checkable behavior: clarification to recovery of a missing site or time, revision to updated operands and windows, timestamp policy to exact or nearest observations, and follow-up requests to quality decisions and evidence. Phase targets, allowed tool calls, and verifier checks are derived from the same source computation. This binding enables phase-specific analysis of agent execution while keeping the underlying telemetry answer traceable.

\subsection{RQ3: Source Grounding and Replayability}
All 532 BTS episodes (356/87/89 train/dev/test) pass contract preflight. Replay rebuilds the tool store, regenerates episodes from frozen inputs, and reproduces released split files byte-for-byte.

Selection replay recovers identities; artifact replay regenerates targets; trajectory replay reproduces scores. Independent raw-source checks find no mismatch in 261 phase-level records across all 89 test episodes (Table~\ref{tab:independent-source-audit}).

\paragraph{Trace-linked failure diagnosis.} We compared final-answer checks with source--contract--tool--trace records for all 216 scored FAIL executions under the released evaluator. The final-answer component passed in 123/216 cases (56.9\%). Final-answer checks captured the complete recorded set of failed obligations in 14/216 cases (6.5\%); in 202/216 (93.5\%), the recorded obligation audit located at least one additional failure in linked records. Linked checks identified substantive phase failures in 181/216 (83.8\%) and execution-level failures in 132/216 (61.1\%). Appendix~\ref{sec:failure-diagnosis-evidence} details the attribution protocol and examples.

\paragraph{Execution-evidence audits.} Across 1,111 scored trajectories, a fixed-question audit compares final answers, full dialogue, and tool records. For 1,103 quality inspections, stream IDs occur verbatim in 106 final answers, 402 dialogues, and all tool-argument records. A fixed rank-to-quality check compares the second requested month's winner with the subsequent quality stream: 41/41 PASS traces match, while 28/40 FAIL traces mismatch (Appendix~\ref{sec:failure-diagnosis-evidence}).

\paragraph{Paid model results.} Each model has three trials per test row: 267 BTS trials and 123 XAI4HEAT trials. Table~\ref{tab:paid-outcomes} reports trial outcomes and $\mathrm{pass}^{3}$ over 89 and 41 row groups, respectively. A group passes when all three trials pass, fails when all trials are scored but at least one fails, and is unresolved when any trial is blocked or unverified. GPT has the highest BTS $\mathrm{pass}^{3}$ count (76/89), followed by Opus (67/89) and Gemini (35/89); on XAI4HEAT, Gemini has 30/41, GPT 27/41, and Opus 26/41.

\begin{table}[!htbp]
\centering\small
\setlength{\tabcolsep}{2.5pt}
\begin{tabular}{llrr}
\toprule
Corpus & Model & Trial P/F/U/B & $\mathrm{pass}^{3}$ P/F/R\\
\midrule
BTS & GPT & 229/30/2/6 & 76/10/3\\
BTS & Gemini & 138/88/12/29 & 35/35/19\\
BTS & Opus & 238/21/0/8 & 67/17/5\\
XAI4HEAT & GPT & 81/42/0/0 & 27/14/0\\
XAI4HEAT & Gemini & 109/13/0/1 & 30/10/1\\
XAI4HEAT & Opus & 94/28/0/1 & 26/14/1\\
\bottomrule
\end{tabular}
\caption{Paid executions evaluated with the released scorer. Trial P/F/U/B: verified PASS, FAIL, UNVERIFIED, BLOCKED/UNSCORED. Group P/F/R: PASS, FAIL, UNRESOLVED. Each model has 267 trials/89 groups on BTS and 123 trials/41 groups on XAI4HEAT.}
\label{tab:paid-outcomes}
\end{table}

\paragraph{Trace review.} The first author examined 87 stratified archived executions, including 14 UNVERIFIED records (Appendix~\ref{sec:human-trace-review}).

\subsection{RQ4: Cross-Corpus Porting to XAI4HEAT}
XAI4HEAT extends the pipeline from BTS graph-linked streams to row-oriented district-heating SCADA. A corpus adapter and two interaction/reporting modules are adapted while runtime, scoring, alignment, and replay are reused. The build reproduces 204 episodes (132/31/41), with preflight validation and source checks for all 41 test rows (Appendix~\ref{sec:xai4heat-portability}).

\section{Conclusion}
BTS-AgentBench compiles telemetry into source-grounded, executable multi-turn evaluation contracts. Frozen-input replay reconstructs 532 BTS and 204 XAI4HEAT episodes, with independent source checks on all 130 test rows. Linked execution records enable phase-specific failure diagnosis. Future work will develop reusable adapters and contracts for systematic cross-domain adaptation.

\section*{Limitations}
\label{sec:limitations}
The BTS replay boundary begins from the frozen normalized catalog, while upstream graph-to-catalog representative selection is fixed in the supplied inputs. Cross-corpus adaptation uses localized interaction and reportability rules alongside the shared runtime and scorer. Ordinary traces expose supporting stream identities; exact evidence-to-result-call lineage is not always separately identified. The benchmark focuses on read-only telemetry analysis, including clarification, revisions, timestamp and quality policies, and evidence-backed reporting.
\section*{Ethical Considerations}

BTS-AgentBench is a read-only offline benchmark over telemetry-derived tasks. It is not intended to authorize, recommend, or evaluate physical control actions in operational buildings or industrial plants. Model traces should therefore be interpreted only as benchmark behavior under a fixed evaluation protocol, not as evidence of deployment safety. Any release or reuse of data should follow the licensing, anonymization, and operational-security requirements of the source dataset.

\section*{Use of Generative AI}

Generative AI tools were used to assist experiments and manuscript revision. The authors remain responsible for methods, results, interpretation, and claims.

\nocite{yao2025taubench,jha2025itbench,patel2026assetopsbench,rayfield2025react}
\bibliography{refs}

\clearpage
\appendix

\section{Author-Led Human Verification}
\label{sec:human-trace-review}

\paragraph{Sampling and materials.} The review panel comprises 87 archived completed trajectories from a 1,125-trajectory inventory, stratified by corpus, model, family, interaction condition and stored outcome. All 14 UNVERIFIED records were included; remaining strata contributed one sampled record each. The resulting panel has 42 PASS, 31 FAIL and 14 UNVERIFIED records. The inventory's 45 BLOCKED/UNSCORED records are not completed review trajectories. Cards expose operator turns, assistant messages, tool schemas and observed tool results, but not model labels, hidden gold or stored evaluator verdicts. Reference contracts do not independently authenticate every historical outbound system payload. Stylistic clues may still reveal model identity. No proportions below are population error-rate estimates.

\begin{table}[!htbp]
\centering\small
\setlength{\tabcolsep}{3pt}
\begin{tabular}{llrrr}
\toprule
Axis & Reviewer & Valid & Invalid & Undetermined \\
\midrule
Semantic & R1 & 86 & 1 & 0 \\
Semantic & R2 & 87 & 0 & 0 \\
Contract & R1 & 0 & 2 & 85 \\
Contract & R2 & 86 & 1 & 0 \\
\bottomrule
\end{tabular}
\caption{Recorded human-review judgments. R2 used a selected-check interpretation of semantic validity; R1 additionally retained unresolved contract-level questions.}
\label{tab:submitted-human-review}
\end{table}

\paragraph{Interpretation.} Semantic agreement is 86/87 (98.85\%), with Cohen's $\kappa=0$: R2 assigns VALID to all records, so observed agreement equals chance agreement under the marginal distributions. Contract agreement is 1/87 (1.15\%), with $\kappa=0.0112$. The reference-contract uncertainty retained by R1 and the narrower checked-item interpretation used by R2 prevent treating these as interchangeable assessments. Thirty of the 31 evaluator FAIL records receive semantic VALID from both reviewers, while neither reviewer pair assigns a shared contract-VALID label. Their differing scopes distinguish the acceptability of selected natural-language answers from satisfaction of complete episode obligations. The review therefore identifies cases for contract-level follow-up, retaining the original UNVERIFIED designations.

\paragraph{Relevance and realism.} Mean obligation-relevance ratings are 4.954/5 (R1) and 4.115/5 (R2); workflow-realism means are 3.954/5 and 4.230/5. The ratings describe the sampled traces under two author-led review protocols with different checking scopes. The first author drew on experience with operational telemetry and factory-management software when interpreting task wording; the recorded scores and review criteria are retained with the audit materials.

\paragraph{Review record provenance.} The review archive retains judgment sheets, source excerpts, and assistant-message request indices. Reviewers inspected original trace messages and assessed them against the available task contracts.

\section{Release and Rebuild Artifacts}

\begin{figure*}[!t]
\centering
\small
\setlength{\tabcolsep}{2pt}
\renewcommand{\arraystretch}{1.15}
\begin{tabularx}{\textwidth}{>{\centering\arraybackslash}Xc>{\centering\arraybackslash}Xc>{\centering\arraybackslash}Xc>{\centering\arraybackslash}Xc>{\centering\arraybackslash}Xc>{\centering\arraybackslash}Xc>{\centering\arraybackslash}X}
\textbf{Raw telemetry} & $\rightarrow$ & \textbf{Telemetry Adapter} & $\rightarrow$ & \textbf{Corpus Contract} & $\rightarrow$ & \textbf{Episode compiler} & $\rightarrow$ & \textbf{Replayable artifact} & $\rightarrow$ & \textbf{Evaluator} & $\rightarrow$ & \textbf{Optional execution} \\
Equipment and point metadata, timestamps, values & & Parse and map records; retain lineage & & Identity, time, tool and policy rules & & Compile dialogue and execute gold operations & & Dialogue, calls, golds, evidence & & Rescore traces and verify contracts & & Agent sees revealed interaction only \\
\end{tabularx}
\vspace{3pt}
\hrule
\vspace{3pt}
\begin{tabularx}{\textwidth}{>{\raggedright\arraybackslash}X>{\raggedright\arraybackslash}X>{\raggedright\arraybackslash}X}
\textbf{Selection replay:} source + config $\rightarrow$ identities & \textbf{Artifact replay:} identity $\rightarrow$ golds & \textbf{Trajectory rescoring:} trace $\rightarrow$ score; no generator feedback \\
\end{tabularx}
\caption{Construction and replay boundaries supplementing the main-text pipeline. Gold construction precedes optional execution.}
\label{fig:lifecycle-boundary}
\end{figure*}

This appendix records the complete release boundary using the same sequence as
the main text: raw telemetry and metadata, read-only tool store, static
executable tasks, typed interaction contracts, operator-facing episodes, and
deterministic scoring targets. The released stage implementations produce the
benchmark artifact. A single-command wrapper binds their inputs, invokes them
in order, and verifies output equality; it contains no construction or scoring
logic. Exact replay starts from checksummed raw archives, the retained
normalized metadata contract, and the fixed 532-row selection identities, then
rebuilds the tool store, static tasks, interaction contracts, phase golds,
evidence, verifiers, and released train/dev/test JSONL.

\begin{table}[t]
\centering
\small
\setlength{\tabcolsep}{4pt}
\renewcommand{\arraystretch}{1.05}
\begin{tabularx}{\linewidth}{>{\raggedright\arraybackslash}p{0.32\linewidth}X}
\toprule
Field & Value \\
\midrule
Released BTS benchmark rows & 532 total rows \\
Released BTS split sizes & train 356 / dev 87 / test 89 \\
Families in the final BTS release & point disambiguation, day mean, relative 24h mean, window mean, window pairwise compare, window rank, timestamp value, timestamp nearest, quality gate \\
Contract preflight & 0 findings under the coded executable-contract checks \\
Exact replay & 2 independent raw-to-episode builds with exact logical and release matches \\
Replay boundary & raw telemetry/metadata $\rightarrow$ tool store $\rightarrow$ static tasks $\rightarrow$ typed contracts $\rightarrow$ operator-facing episodes $\rightarrow$ scored release \\
\bottomrule
\end{tabularx}
\caption{Released BTS artifact snapshot used by the paper.}
\label{tab:appendix-release-snapshot}
\end{table}

\section{Contract Transformation Details}
\label{sec:contract-transformation-details}
\paragraph{Deterministic contract transformation.}
Typed contract composition and repair denotes an ordered family of contract transformations: some compose a new interaction obligation, while others restore alignment among a turn, its tool call, gold fields, evidence, and verifier. For a normalized read-only store $D$, stage $k$ is
\begin{equation}
\begin{aligned}
z_k &= \operatorname{Exec}_D(a_k(C)),\\
T_k(C;D) &={}
\begin{cases}
U_k(C,z_k), & P_k(C)=1,\\
C, & P_k(C)=0.
\end{cases}
\end{aligned}
\end{equation}
where $P_k$ is a typed eligibility predicate, $a_k$ constructs tool arguments from existing contract fields, and $U_k$ performs the coupled update. Predicates inspect fields such as task family, phase types, available stream IDs, and exact-match status; they do not classify free-form text. If a predicate holds, the stage executes the declared read-only operation and atomically updates the affected operator turn, phase sequence, canonical calls, gold fields, evidence, milestones, and verifier. A surface-only stage may normalize wording but cannot change telemetry fields. The stages run in a fixed order, and a false predicate records a no-op. Static ID, split, source family, and source backlink remain fixed throughout.

Each applied stage must satisfy three coupled conditions. First, source preservation keeps all unaffected static phases, stream bindings, and split fields unchanged. Second, execution grounding requires every newly introduced value, timestamp, aggregate, or quality statistic to equal $z_k$ or a declared deterministic policy function over prior phase state. Third, discourse alignment requires the rendered turn, its referenced prior state, required tool path, gold fields, evidence, and verifier to describe the same operation. The coupled update $U_k$ writes these fields together, and contract preflight independently re-executes and checks them before retention.

\paragraph{Executable example.}
For the timestamp row in Table~\ref{tab:episode-compilation}, the seed contract contains one exact-timestamp phase grounded in value 12.9457 at \texttt{2022-02-03T07:03:23.640Z}. Clarification composition marks time as a missing slot and renders a request for it. The timestamp-revision stage constructs the public-time request at 07:03, executes an exact probe followed by the permitted nearest fallback, and appends the returned timestamp, value, fallback status, and 23.64-second offset as a second phase. Multi-axis composition executes the quality tool for the containing week and appends coverage 1.0, gap ratio 1.0563, and decision \texttt{answer}. Finally, the reporting rule combines the typed nearest-reading and quality states to derive action \texttt{answer} with reason \texttt{nearest\_but\_acceptable}. Telemetry facts come from tool execution and policy labels from finite declared rules. Appendix~\ref{sec:exact-row-replay} gives the complete raw-to-score trace.

\paragraph{Audit trail and replay.}
Every stage appends its name, status, and before/after contract summaries to an ordered history. Before episode lifting, the static audit recomputes point resolution and family outputs against the tool store and referenced raw history. After lifting, a separate contract audit checks phase/turn cardinality and required fields; renders each stored gold as an answer and sends it through the same scorer; re-executes timestamp and quality operations; recomputes derived preference and commitment fields; and checks evidence and prompt--phase alignment. A row is retained only if these checks accept the coupled contract. With the store, selected static identity, stage order, predicates, and templates fixed, reconstruction yields the same turns, calls, phase golds, evidence, verifier, and history; Section~\ref{sec:artifact-validation} reports exact equality across independent builds.

The retained histories expose how often each typed stage changes a contract: timestamp policy 120 rows, nearest wording 60, timestamp-value revision 60, clarification alignment 60, goal revision 293, alternate-point revision 60, multi-axis composition 472, and terminal reporting commitment 532. Counts overlap because one episode may receive several construction stages; Table~\ref{tab:repair-profile} in the appendix reports the complete profile.

\section{Interaction-Structure Details}
\label{sec:interaction-structure-details}
Static-only, single-turn, state-free, and full-episode projections show which facts are introduced by dialogue, carried across turns, revised, checked against policy, and attributed to source evidence. Table~\ref{tab:structural-contract-ablation} summarizes these structural projections. The projections identify which obligations enter at each compilation stage.

\begin{table}[t]
\centering\small
\setlength{\tabcolsep}{2pt}
\begin{tabularx}{\columnwidth}{>{\raggedright\arraybackslash}Xrrrrr}
\toprule
Family & Rows & Clarify & Revision & Quality & Evidence \\
\midrule
Point disambiguation & 10 & 10 & 10 & 10 & 10 \\
Day mean & 10 & 10 & 10 & 10 & 10 \\
Relative 24h mean & 10 & 10 & 10 & 10 & 10 \\
Window mean & 10 & 10 & 10 & 10 & 10 \\
Pairwise compare & 10 & 10 & 10 & 10 & 10 \\
Window rank & 10 & 10 & 10 & 10 & 10 \\
Timestamp value & 10 & 10 & 10 & 10 & 10 \\
Timestamp nearest & 10 & 0 & 10 & 10 & 10 \\
Quality gate & 9 & 9 & 9 & 9 & 9 \\
\bottomrule
\end{tabularx}
\caption{Interaction obligations encoded in the final test contracts. Explicit reporting/policy branches occur in 54/89 rows; clarification is absent in nearest-timestamp rows by construction.}
\label{tab:final-interaction-obligations}
\end{table}

\begin{table}[t]
\centering\small
\begin{tabularx}{\columnwidth}{lX}
\toprule
Contract view & Obligations retained or removed on the 89-row test panel \\
\midrule
Static source task & Source identity and base query; no generated dialogue-state obligation. \\
Single-turn task & Base operation/result; omits 79 clarification slots, 304 revision turns, and 98 post-answer turns. \\
Multi-turn without revisions & Dialogue remains, but removes 89 row-level revision obligations and downstream state-update checks. \\
Multi-turn without policy branch & Removes explicit reporting/policy branches in 54 rows. \\
Full compiled episode & Clarification where required, revisions, quality/reporting, commitment, and evidence contracts. \\
\bottomrule
\end{tabularx}
\caption{Structural contract projections of the 89-row test panel.}
\label{tab:structural-contract-ablation}
\end{table}

\section{Local Construction Dependency Audit}
\label{sec:locality-audit}
The compiler, source adapter, DuckDB tool runtime, episode builders, replay machinery, and deterministic evaluator run as local Python/SQL programs. Static dependency inspection places external model-client construction only in the optional frontier-evaluation runner. Benchmark construction, replay, and deterministic scoring execute locally after the source corpora are acquired.

The stage-level dependency audit is reported in Appendix~\ref{sec:locality-audit}.

\begin{table}[t]
\centering\small
\setlength{\tabcolsep}{3pt}
\begin{tabularx}{\columnwidth}{>{\raggedright\arraybackslash}p{0.29\columnwidth}>{\centering\arraybackslash}p{0.14\columnwidth}>{\centering\arraybackslash}p{0.17\columnwidth}>{\raggedright\arraybackslash}X}
\toprule
Stage & Raw corpus & External LLM & Local execution \\
\midrule
Parse and normalize & Yes & No & Yes \\
Tool store and task generation & Yes & No & Yes \\
Episode and gold compilation & Yes & No & Yes \\
Deterministic replay/scoring & Yes & No & Yes \\
Optional frontier evaluation & No & Optional & Separate from benchmark construction \\
\bottomrule
\end{tabularx}
\caption{Dependency audit after source acquisition. Construction, replay, and deterministic scoring use local Python/SQL components.}
\label{tab:locality-audit-table}
\end{table}

\section{Evaluator Observability Validation}
\label{sec:observability-validation}
\paragraph{Compact results summary.}
\label{sec:evaluator-observability}
A secondary audit measures direct task-contract coverage: the fraction of mandatory applicable task-contract instances observed by each benchmark's primary score. Coverage is 93.4\% for BTS-AgentBench, 17.9\% for $\tau^3$ Airline, and 78.0\% for ToolSandbox Base129. Six paired official-evaluator traces further show representative cases where task-facing behavior changes while the primary score remains unchanged. Full obligation mappings, reward-component comparisons, mutation/control validation, and counterexamples are reported in Appendix~\ref{sec:observability-validation}.

\paragraph{Main-text audit formulation and summary.}
We define task-contract coverage as the fraction of mandatory applicable task-contract instances directly observed by the benchmark's reported primary score. Obligations are extracted from each benchmark's own task-facing contract and mapped to the signals consumed by its primary evaluator. This yields a common denominator principle while preserving each benchmark's native obligation taxonomy.

The full cross-benchmark coverage table is reported in Appendix~\ref{sec:observability-validation}.

The final BTS test contract reaches 93.4\% direct task-contract coverage; $\tau^3$ Airline and ToolSandbox reach 17.9\% and 78.0\%, respectively. The external evaluators fully cover their explicit reward components, and six paired official-evaluator traces confirm representative cases where task-facing behavior changes while the primary score remains unchanged. Full obligation mappings, mutation/control validation, and case-level counterexamples are reported in Appendix~\ref{sec:observability-validation}.

\begin{table}[t]
\centering\scriptsize
\setlength{\tabcolsep}{3pt}
\begin{tabularx}{\columnwidth}{>{\raggedright\arraybackslash}Xrr>{\raggedright\arraybackslash}p{0.23\columnwidth}}
\toprule
Benchmark & Applicable & Direct (coverage) & Fully/partial/not direct \\
\midrule
BTS-AgentBench (27 types) & 1,352 & 1,263 (93.4\%) & 26/0/1 \\
$\tau^3$ Airline (6 types) & 112 & 20 (17.9\%) & 0/1/5 \\
ToolSandbox Base129 (5 types) & 268 & 209 (78.0\%) & 2/2/1 \\
\bottomrule
\end{tabularx}
\caption{Within-benchmark direct task-contract coverage. Denominators and taxonomies are benchmark-specific; percentages do not compare evaluator accuracy or construct validity.}
\label{tab:external-task-contract-coverage}
\end{table}

\paragraph{Coverage and counterexample details.}
Canonical positives and designed semantic mutations test specific contract branches. Their acceptance/detection counts are not estimates of evaluator precision or recall on naturally generated trajectories. Independent judgments of actual valid and invalid model behavior would be needed for that claim; uncovered representation variants and ambiguity cannot be ruled out by these tests alone.
For the final test contract, the ordinary-trace scorer directly enforces 1,263/1,352 mandatory applicable instances (93.4\%). Of 27 mandatory obligation types, 26 are fully direct, none is partially direct, and one is not direct. Protocol completion is directly checked on all applicable test rows using the declared protocol graph and observable execution events. The sole non-direct type is exact evidence-to-result provenance: ordinary traces expose stream membership but not a unique originating call/result for each evidence item. The canonical gate is 89/89 PASS; the actual-format suite detects 446/446 intended component violations and accepts 104/104 controls. The dedicated protocol suite separately detects 419/419 violations and accepts 10/10 valid controls; these suites overlap in scope and are not summed.

We applied the same denominator principle to the pinned $\tau^3$ Airline and ToolSandbox Base129 contracts. Task-facing obligations were extracted in a separate code path before evaluator mapping and then frozen. $\tau^3$ Airline (50 tasks) yields 112 mandatory applicable instances, of which 20 are directly primary-scored (17.9\%); type status is 0 fully direct, 1 partially direct, and 5 not direct. ToolSandbox 0.0.1 contributes 129 base scenarios and 268 mandatory applicable instances, of which 209 are direct (78.0\%); type status is 2 fully direct, 2 partially direct, and 1 not-direct type.

The same audit also recovers complete coverage of the explicit reward components used by the external evaluators: 56/56 (100\%) for $\tau^3$ Airline and 336/336 (100\%) for ToolSandbox. Comparing these numbers with task-contract coverage shows that reward-component coverage and task-facing observability measure different properties.

We validated representative observability gaps with six deterministic paired traces run through the pinned official primary evaluators. Three $\tau^3$ pairs (tasks 0, 3, and 6) vary cancellation-refusal behavior, membership verification, or insurance-refusal behavior while preserving the reward inputs; each pair scores 1.0/1.0. Three ToolSandbox insufficient-information pairs vary clarification versus unsupported answers while preserving the scored minefield condition; each pair scores 1/1. Manual inspection confirms that the task-facing obligation changes while the evaluator-observed primary condition remains fixed. These pairs provide execution-level witnesses for the distinction between reward-component coverage and task-contract observability.

The comparison uses a shared denominator principle with benchmark-specific obligation taxonomies. ToolSandbox's natural-language goals require component-level semantic mapping, while $\tau^3$ exposes policy obligations through its task and policy contracts. We use the resulting percentages to compare contract-to-evaluator alignment.

\begin{table}[t]
\centering\scriptsize
\setlength{\tabcolsep}{3pt}
\begin{tabularx}{\columnwidth}{>{\raggedright\arraybackslash}p{0.25\columnwidth}>{\raggedright\arraybackslash}Xcc}
\toprule
Benchmark / case & Varied task-facing behavior & Compliant & Violating \\
\midrule
$\tau^3$ / task 0 & Refuse / agree to ineligible cancellation & 1.0 & 1.0 \\
$\tau^3$ / task 3 & Verify membership / omit profile lookup & 1.0 & 1.0 \\
$\tau^3$ / task 6 & Refuse / offer disallowed insurance & 1.0 & 1.0 \\
ToolSandbox / city & Ask for location / assert unsupported city & 1 & 1 \\
ToolSandbox / holiday & Ask which date / assert unsupported interval & 1 & 1 \\
ToolSandbox / distance & Ask for location / assert unsupported distance & 1 & 1 \\
\bottomrule
\end{tabularx}
\caption{Six selected official-evaluator counterexamples. Each pair preserves the evaluator-observed primary condition while changing the stated task-facing behavior.}
\label{tab:external-counterexamples}
\end{table}

The final deterministic scorer accepts all 89 canonical execution traces. Across the actual-format suite it detects 446/446 intended component violations and accepts 104/104 valid controls. A dedicated protocol suite separately detects 419/419 protocol violations and accepts 10/10 valid alternative orderings. These suites overlap and are reported separately. Repeated scoring produces identical serialized outputs. 

The final mandatory task-contract coverage is 1,263/1,352 instances (93.4\%): 26/27 types are fully direct, zero are partially direct, and one is not direct. Protocol completion is directly checked on every applicable test row. Exact evidence-to-result provenance remains a documented boundary because ordinary traces expose stream membership but not a unique source call/result for each evidence item.

\subsection{Reference Algorithm: Raw Telemetry to Replayable Gold}
The following schematic summarizes the source-to-episode compiler used in construction.

\begin{lstlisting}[style=methodpython,caption={Source-to-episode compiler (schematic).}]
for source_stream in normalized_catalog(raw_metadata):
    history = open_raw_history(source_stream)
    materialize_read_only_indices(history)

for candidate in enumerate_family_candidates(tool_store):
    if not eligibility_contract(candidate):
        continue
    static = execute_source_task(candidate, tool_store)
    # static now binds source identity, time/window, tool result, evidence

    episode = seed_interaction_contract(static)
    for transform in typed_episode_transforms(static.family):
        episode = transform(episode)
        # Any new telemetry-dependent target is executed, not invented.
        episode = execute_and_bind_new_gold(episode, tool_store)

    validate_contract_against_source(episode, tool_store)
    emit_selected(episode)  # frozen identity selection

# Later and separately:
# model/controller -> archived trace
# offline evaluator(episode, trace)
\end{lstlisting}

The compiler never observes evaluated model responses, and evaluated agents do not receive the hidden contract fields used to construct or score their tasks.

\subsection{Raw-to-Static Compilation Details}

The raw-to-static compiler starts from processed BTS metadata and the raw
stream archives. It normalizes site, point, equipment, and location text;
builds a point inventory; records ambiguity candidates; and scans each matched
raw stream to materialize timestamp coverage summaries. Only metadata points
that can be matched to raw histories are promoted to \emph{tool-ready points}.
From those matched points, the compiler materializes the DuckDB-backed tool
store used by the benchmark runtime: a raw stream index, per-stream quality
statistics, daily/weekly/monthly aggregates, calendar profiles, and previews.

Family builders enumerate static executable tasks over this store using the
eligibility rules in Section~\ref{sec:static-layer}. BTS\_C is held out for test; non-held-out
candidates are deterministically assigned to train and development under
point-class and temporal diversity caps. The retained identity contract fixes
the evaluated 532 candidates after recomputation and fails if an identity is
missing or duplicated. It stores no generated conversation or model output.

\subsection{Static-to-Episode Contract Derivation}

The episode builder derives each conversation from a
bounded typed interaction contract from the retained static task. The builder
first selects an interaction mode from the task family and the static-task
surface, including
direct-answer, clarify-time, implicit-nearest, and quality-rationale modes,
plus optional site clarification. That mode determines which information is
withheld initially, which clarification slots must be filled, and which
post-answer turns are required later in the exchange.

Each converted row therefore adds deterministic fields such as an initial user
message, hidden user instruction, required clarification slots, clarification
questions, clarification answers, post-answer user turns, termination reply,
interaction verifier, and interaction milestones. At the same time, it carries
forward the static row's telemetry computation, acceptable tool-call sets, gold
final answer, evidence bundle, verifier, difficulty proxy, and metadata. Later
typed contract composition and family-specific transformations may add more explicit
phase structure, but every operator-facing episode continues to point back to
the same static executable task.

\begin{lstlisting}[style=traceblock,caption={Public replay with the raw archive directory as the only user-selected data path.}]
make replay RAW_DIR=/absolute/path/to/BTS_RAW_ARCHIVES

# Equivalent expanded command:
python scripts/replay_release.py \
  --raw-dir /absolute/path/to/BTS_RAW_ARCHIVES \
  --work-dir ./data/local-build/release-replay \
  --raw-runs 2 \
  --controller-audit
\end{lstlisting}

The raw directory must contain \path{Site_Aaa.zip}, \path{Site_Baa.zip}, and
\path{Site_Caa.zip}. It is the only input path supplied by the caller. The
repository contains the checksummed normalized catalog, source metadata, fixed
532-row selection contract, and expected release manifest. The replay creates
the DuckDB tool stores and every intermediate under the work directory; it does
not consume a pre-existing local database.

The recorded environment is Python 3.11.11, DuckDB 1.5.0, NumPy 1.26.4,
pandas 3.0.1, PyArrow 23.0.1, and RDFLib 7.6.0. Two independently rebuilt
stores match all 11 sorted logical exports, and both complete builds reproduce
all 356/87/89 rows as exact JSON-object and file matches.

\begin{table}[t]
\centering
\scriptsize
\setlength{\tabcolsep}{3pt}
\renewcommand{\arraystretch}{1.04}
\begin{tabularx}{\columnwidth}{>{\raggedright\arraybackslash}p{0.25\columnwidth}>{\raggedright\arraybackslash}p{0.35\columnwidth}X}
\toprule
Check & Verifies & Boundary \\
\midrule
Input checksum & Exact raw and retained files & Not upstream collection validity \\
Selection identity & Unique regenerated retained candidates & Not random representativeness \\
Static/final equality & Every released object and byte & Not provider repeatability \\
Contract preflight & Schema, gold, evidence, verifier, runtime alignment & Not every possible annotation error \\

\bottomrule
\end{tabularx}
\caption{Scope of the deterministic validation checks.}
\label{tab:validation-matrix}
\end{table}

\begin{table}[t]
\centering
\scriptsize
\setlength{\tabcolsep}{3pt}
\renewcommand{\arraystretch}{1.03}
\begin{tabularx}{\columnwidth}{>{\raggedright\arraybackslash}Xr}
\toprule
Typed construction stage & Rows modified \\
\midrule
Timestamp policy / nearest wording & 120 / 60 \\
Timestamp-value / alternate-point revision & 60 / 60 \\
Clarification / goal revision & 60 / 293 \\
Multi-axis composition & 472 \\
Terminal reporting commitment & 532 \\
Controller-proxy transformation & 0 \\
\bottomrule
\end{tabularx}
\caption{Overlapping typed-stage applications. Counts show how often each typed construction stage is applied.}
\label{tab:repair-profile}
\end{table}

\subsection{Exact Row-Level Replay}
\label{sec:exact-row-replay}

This example traces a source observation through the final episode construction and replay workflow.

For \path{test_timestamp_value_lookup_00051}, raw member
\path{Site_Caa/2254.pickle} contains 194,563 observations for stream
\path{c24589e8_a1f3_4529_b409_5a56761c9d20}; the selected record is
\texttt{2022-02-03T07:03:23.640Z = 12.9457}. Fresh preprocessing reconstructs
the point and observation tables. The static builder emits
\texttt{resolve\_point} followed by exact \texttt{lookup\_observation}, with
that value and stream as gold and evidence.

Episode construction withholds the timestamp, supplies it through deterministic
clarification, revises the request to nearest lookup at 07:03, checks the week
beginning January 31, and requests a reporting commitment and evidence. The
recomputed phase targets are: exact value 12.9457; nearest offset 23.64 seconds;
quality action \texttt{answer} with coverage 1.0 and gap ratio 1.0563; and final
reason \texttt{nearest\_but\_acceptable}. The reconstructed row retains its source stream and phase targets, and the rebuilt test artifact matches the released 89-row file.

\subsection{Final-Artifact Worked Episode}
\label{sec:final-worked-episode}
\label{sec:final-compilation-detail}

\begin{table*}[!t]
\centering
\small
\setlength{\tabcolsep}{4pt}
\renewcommand{\arraystretch}{1.07}
\begin{tabularx}{\textwidth}{>{\raggedright\arraybackslash}p{0.18\textwidth}>{\raggedright\arraybackslash}X}
\toprule
Compilation stage & Deterministic rule and emitted artifact \\
\midrule
Raw observation & \path{Site_Caa.zip/Site_Caa/2254.pickle} contains stream \path{c24589e8_a1f3_4529_b409_5a56761c9d20}; its value at \texttt{2022-02-03T07:03:23.640Z} is $12.9457$. \\
Adapter output & The stream maps to \path{Air_Differential_Pressure_Sensor} on \texttt{BTS\_C Zone 005}; its source value remains unitless. \\
Corpus Contract & Stable stream identity; UTC timestamps; half-open calendar windows; deterministic exact/nearest lookup and quality operations; no engineering unit without authoritative source metadata. \\
Selected static task & \texttt{timestamp\_value\_lookup}; resolve Zone 005 and query the source timestamp, preserving stream, value, and evidence. \\
Mode and visibility & The missing-time request selects \texttt{clarify\_time}; the simulator supplies ``07:03:23.64 UTC on February 3, 2022.'' \\
Turns and state updates & Revise the request to the nearest reading around 07:03, assess the week beginning January 31, then decide whether to report, abstain, or re-clarify. \\
Tool results & Exact and nearest value $12.9457$; nearest offset $23.64$ seconds; weekly coverage $1.0$, gap ratio $1.0563$, quality decision \texttt{answer}. \\
Gold/evaluator & Final action \texttt{answer}, reason \texttt{nearest\_but\_acceptable}; source-stream evidence. Phase values/times, reporting, stream identity, and protocol are scored separately. \\
\bottomrule
\end{tabularx}
\caption{Field-level compilation of \texttt{test\_timestamp\_value\_lookup\_00051}, complementing the main-text construction diagram.}
\label{tab:episode-compilation}
\end{table*}
The following trace makes each transformation explicit for the final artifact's \path{test\_timestamp\_value\_lookup\_00051}. The source metadata maps \path{c24589e8_a1f3_4529_b409_5a56761c9d20} to point class \path{Air_Differential_Pressure_Sensor} on BTS\_C Zone 005. Its raw archive member contains 194,563 observations; the selected record is \texttt{2022-02-03T07:03:23.640Z, 12.9457}. Values are unitless raw source values because the catalog has no authoritative engineering unit.

\begin{table}[t]
\centering
\scriptsize
\setlength{\tabcolsep}{3pt}
\renewcommand{\arraystretch}{1.04}
\begin{tabularx}{\columnwidth}{>{\raggedright\arraybackslash}p{0.21\columnwidth}>{\raggedright\arraybackslash}X}
\toprule
Stage & Source-derived contract and expected state/result \\
\midrule
Static task & Resolve BTS\_C / Zone 005 / air differential pressure; query the exact timestamp 2022-02-03 07:03:23.640 UTC. \\
Initial dialogue & ``What was the air differential pressure reading on Zone 005 in BTS\_C?'' The timestamp is missing, so ask the operator; the revealed clarification supplies it. \\
Initial execution & \texttt{resolve\_point} returns exactly one stream. Exact \texttt{lookup\_observation} returns raw value 12.9457 at the requested timestamp. \\
Revision & ``Keep the same signal and site ... around 07:03 ... give me the nearest available reading.'' Retain site, entity, point class, and stream; replace exact time with the 07:03 probe and nearest policy. \\
Nearest execution & Exact probe at 07:03:00 has no match. Nearest lookup returns the 07:03:23.640 observation, value 12.9457, offset 23.64 seconds; cadence-derived tolerance is 299.761 seconds, so status is acceptable. \\
Quality follow-up & The operator asks about the week beginning January 31. Retain the resolved stream; add the half-open UTC week [2022-01-31, 2022-02-07). Quality output is coverage 1.0, gap ratio 1.0563, decision \texttt{answer}. \\
Result-conditioned final & Given the nearest assessment and quality result, the reporting rule derives \texttt{answer / nearest\_but\_acceptable}; the evidence target remains the resolved stream. \\
\bottomrule
\end{tabularx}
\caption{Worked final-artifact compilation: source identity and observation, dialogue revision, deterministic tool results, and derived reporting/evidence golds.}
\label{tab:final-worked-episode}
\end{table}

The full operator exchange and gold responses are:
\begin{lstlisting}[style=traceblock,caption={Final-artifact dialogue and canonical result chain for the worked row.}]
Operator: Operator handoff: "What was the air differential pressure reading on Zone 005 in BTS_C?" Use the building tools and ask me for any missing site or time detail before querying.
Expected agent: Which date, week, month, time window, or exact timestamp should I use?
Operator: I mean 07:03:23.64 UTC on February 3, 2022.
Expected agent: The exact recorded raw value at 2022-02-03 07:03:23.640000 UTC was 12.9457.
Operator: Now keep the same signal and site, but if I only know it was around 07:03 UTC on February 3, 2022, give me the nearest available reading.
Expected agent: No exact observation was available at 2022-02-03 07:03:00 UTC. The nearest recorded raw value was 12.9457 at 2022-02-03 07:03:23.640000 UTC (23.64 seconds away).
Operator: For the week beginning January 31, 2022, would you answer or abstain based on data quality?
Expected agent: I would answer. Reason: healthy. Observed coverage: 1.0. Gap ratio: 1.0563.
Operator: Given the public-time reading and the data-quality check, should I report it as-is, abstain, or ask for more time detail before reporting it?
Expected agent: Report it as-is. Reason: nearest but acceptable.
Operator: Which stream or point did you base that on?
Expected agent: I based this on the air differential pressure sensor point on BTS_C Zone 005.

Gold trajectory:
  resolve_point(BTS_C, Zone 005, Air_Differential_Pressure_Sensor)
    -> unique stream c24589e8...9d20
  lookup_observation(stream, 07:03:23.640, exact)
    -> 12.9457, exact_match=true
  lookup_observation(stream, 07:03:00, exact)
    -> no exact observation
  lookup_observation(stream, 07:03:00, nearest)
    -> 12.9457 at 07:03:23.640; offset=23.64 s;
       median_step=599.522 s; tolerance=299.761 s; acceptable
  inspect_quality_window(stream, [2022-01-31, 2022-02-07), week)
    -> observed_fraction=1.0; gap_ratio=1.0563
  derive_reportability(nearest_assessment, quality_result)
    -> answer / nearest_but_acceptable
  evidence -> c24589e8...9d20
\end{lstlisting}
The evaluator checks resolved identity, exact and nearest timestamps/value, offset and acceptability, quality fields, final commitment, evidence identity, and protocol independently. The final commitment depends on the actual nearest assessment and quality output; equality with a stored answer alone is not its provenance.

\begin{table}[!htbp]
\centering\small
\setlength{\tabcolsep}{3pt}
\begin{tabularx}{\columnwidth}{>{\raggedright\arraybackslash}Xrr}
\toprule
Independent source check & Records & Matches \\
\midrule
Single-stream means & 80 & 80 \\
Pairwise comparisons (40 mean operands) & 20 & 20 \\
Exact/nearest observations & 72 & 72 \\
Window quality statistics & 89 & 89 \\
\midrule
Total phase records & 261 & 261 \\
\bottomrule
\end{tabularx}
\caption{Independent raw-source checks across all 89 test episodes, with zero mismatches. Means use UTC half-open windows; numeric statistics match at four decimals, timestamps exactly, and nearest offsets within 1 ms. Archive lookup uses the frozen raw index.}
\label{tab:independent-source-audit}
\end{table}

\subsection{Independent Raw-Source and Trace Replay Checks}
The three replay levels recover selected identities, regenerate source-derived targets, and score stored executions. The independent raw-source audit covers all 89 held-out test episodes with 261 phase-level records: 80 single-stream means, 20 pairwise records (40 mean operands), 72 timestamp records, and 89 quality windows. All match under stated tolerances. Thus 261 counts check records, not episodes or individual scalar computations; train/dev are not independently audited here.

A fixed synthetic assistant-action sequence produces identical user-simulator outputs on two runs for all 89 test rows. Together, these checks cover source computation, tool execution, simulator behavior, artifact reconstruction, and scoring determinism.

\label{sec:independent-replay-checks}
The local audit runner \path{scripts/run_methodology_evidence_audit.py} reads raw pickle payloads across 40 streams without calling production aggregation or quality functions. Means and quality statistics match at four decimal places, timestamps match exactly, and nearest offsets agree within 1 ms. The frozen raw index identifies archive members, so catalog identity selection and policy thresholds are not independently validated.

The observable evaluator produces byte-identical rescoring twice (canonical JSONL SHA-256: \shahex{3ad0b8c5c857695564a5cc5a0d27e3c98a11e4d660c62c3ee06763bcf440c9b9}). A deterministic synthetic assistant-action sequence also yields identical simulator outputs on two runs for all 89 test rows and 659 transitions. This test isolates the determinism of user-turn exposure under a fixed action sequence.

The audit summary and per-check records are under \path{artifacts/methodology-evidence-v1/}; its hashes identify that audit snapshot, not final protocol-augmented splits. Final split and scorer-output hashes are recorded in the final artifact's revision, repeatability, and checksum manifests. 

\FloatBarrier
\section{Failure-Diagnosis Evidence Ablation}
\label{sec:failure-diagnosis-evidence}
\paragraph{Comparison protocol.}
We examined all 216 scored FAIL trajectories under the released evaluator using archived executions. For each trajectory, we deduplicated repeated issue strings and merged the core, final and last-phase issues when they described the same answer fact. An obligation was classified as final-answer-visible when it belonged to the final-answer check or last public phase. Earlier or missing phases, interaction obligations, tool-call arguments and cardinality, stream identity, state carryover, temporal checks, evidence and protocol completion were classified as linked-record obligations. An answer-only diagnosis was complete when it identified at least one failed obligation and no additional linked-record obligation remained. The answer-only condition was credited with the evaluator's final-answer issues.

\begin{table}[t]
\centering\small
\setlength{\tabcolsep}{4pt}
\begin{tabular}{p{0.67\columnwidth}rr}
\toprule
Recorded observation & Count & Rate \\
\midrule
Final-answer component has a failure & 93 & 43.1\% \\
Final-answer component passes & 123 & 56.9\% \\
Answer-only diagnosis complete & 14 & 6.5\% \\
Linked record adds failure information & 202 & 93.5\% \\
Substantive phase failure & 181 & 83.8\% \\
Execution-check failure & 132 & 61.1\% \\
\bottomrule
\end{tabular}
\caption{Failure observability for 216 scored FAIL trajectories.}
\label{tab:failure-diagnosis-summary}
\end{table}

\paragraph{Fixed-question observation protocol.}
The archive contains 1,125 completed traces among 1,170 planned identities. We analyzed the 1,111 scored PASS/FAIL trajectories, retaining the same eligible population within each question. View A contains the final public assistant response; view B contains all public assistant responses; view C contains recorded tool arguments and results. We asked for the exact quality-call stream identifier and time-window endpoints, and the observation-lookup stream identifier, returned value, and observed timestamp. In the text views, an item was marked explicit when its identifier, timestamp, or numeric literal matched the recorded field. This procedure measures the recoverability of concrete execution fields across views.

\begin{table}[t]
\centering\small
\setlength{\tabcolsep}{3pt}
\begin{tabular}{p{0.43\columnwidth}rrrr}
\toprule
Fixed query & Eligible & A & B & C \\
\midrule
Quality stream ID & 1103 & 106 & 402 & 1103 \\
Quality window endpoints & 1103 & 0 & 0 & 1103 \\
Lookup stream ID & 774 & 84 & 250 & 774 \\
Lookup value literal & 774 & 391 & 488 & 774 \\
Lookup timestamp & 774 & 23 & 76 & 774 \\
\bottomrule
\end{tabular}
\caption{Recorded execution fields recoverable from the final answer (A), all public assistant responses (B), or tool records (C). The question and denominator are fixed within each row; text retrieval uses exact literal matches.}
\label{tab:fixed-question-info}
\end{table}

\paragraph{Contract-to-trace evidence matrix.}
We joined each of the 216 archived FAIL trajectories to its source episode contract using corpus, scenario, and canonical trial slot. The joined matrix spans 130 source contract definitions and records final gold fields, allowed tool-call sets, expected and observed tool types, observed stream arguments, and the trace location. It assigns an executable compliance verdict for the rank-to-quality invariant when that invariant is observable. For the other families, the matrix retains the relevant contract and trace evidence for obligation-specific inspection. This yields an auditable, row-level linkage between the 216 aggregate outcomes and concrete execution evidence.

\paragraph{One fixed obligation across ranking executions.}
The executable requirement is to carry the winner of the \emph{second requested} month into the quality inspection. We selected 81 archived trajectories with two \texttt{rank\_window} results followed by an \texttt{inspect\_quality\_window} call, and compared the second result's first-ranked \texttt{stream\_id} to the quality call's \texttt{stream\_id}. The 41 overall-PASS executions all matched. Among 40 overall-FAIL executions, 12 matched and 28 differed. For example, Slot \texttt{150} revised March to February: its March winner was \texttt{1d68b345...}, its February winner was \texttt{6b843e55...}, and its quality call used \texttt{1d68b345...}. The tool-result/argument comparison identifies the precise stream-selection mismatch under a single unchanged requirement. The eligibility set consists of trajectories with the required observed calls, and the reference is the archived ranking result.

\paragraph{Execution checks.}
The linked records identify protocol-completion failures in 64 trajectories, material tool-argument failures in 43, stream-identity failures in 43, evidence-provenance failures in 32, and state-carryover failures in 23. The categories overlap. Of the 93 executions with a failing final-answer component, 14 have a complete recorded failure explanation at the final-answer level and 79 have additional linked failures. The other 123 have a passing final-answer component.

\paragraph{Representative archived executions.}
Slot \texttt{150} (BTS ranking, GPT) passes the final-answer check, but its quality query uses stream \texttt{1d68...} rather than the active winner stream \texttt{6b843...}. Slot \texttt{275} (BTS point disambiguation, Gemini) passes the final report/abstain check but omits the required evidence follow-up. Slot \texttt{515} (BTS quality gate, Gemini) gives the final abstention rationale while leaving the required earlier quality-preference decision unanswered. Slot \texttt{648} (BTS window mean, Opus) passes the final-answer component while propagating an incorrectly resolved stream through aggregation, observation lookup and quality inspection. These examples connect final-response judgments to recorded intermediate obligations and tool results.

\paragraph{Reproduction.}
The archived analysis is specified by \path{experiments/provenance_diagnostic_value/run.py} and the recorded summary and per-failure CSV. It uses 1,125 completed scorer objects indexed against 1,170 trial identities and selects 216 final FAIL cases. The per-failure export records deduplicated answer-visible and linked obligations, final-component status, and trace/score hashes. Two executions generated byte-identical outputs; the source audit includes SHA-256 hashes for the input reconciliation, released scorer bundle, analysis script and result files. The failure categories describe the released evaluator's recorded obligation checks.

\subsection{Archived Task-Family Outcomes}
\label{sec:family-performance}
We grouped the 1,111 scored PASS/FAIL trial identities by their compiled task family. Each entry below is an observed trajectory outcome under its released episode contract.
\begin{table}[t]
\centering\small
\setlength{\tabcolsep}{3pt}
\begin{tabular}{lrrr}
\toprule
Task family & Scored & FAIL & FAIL (\%) \\
\midrule
Day mean & 170 & 7 & 4.1 \\
Relative 24h mean & 179 & 13 & 7.3 \\
Window mean & 149 & 6 & 4.0 \\
Point disambiguation & 79 & 9 & 11.4 \\
Nearest timestamp & 143 & 38 & 26.6 \\
Exact timestamp value & 145 & 50 & 34.5 \\
Pairwise comparison & 90 & 28 & 31.1 \\
Window ranking & 81 & 40 & 49.4 \\
Quality-aware reporting & 75 & 25 & 33.3 \\
\midrule
Total & 1111 & 216 & 19.4 \\
\bottomrule
\end{tabular}
\caption{Observed full-trajectory outcomes by task family across the scored paid-run archive.}
\label{tab:family-scored-outcomes}
\end{table}

\section{Additional Discussion}
\label{sec:additional-discussion}

The central result is a deterministic construction methodology that keeps the source-to-score path executable. A Telemetry Adapter maps corpus-specific records into a shared representation; a Corpus Contract makes identity, temporal, missing-data, and tool semantics explicit; and a deterministic compiler turns selected source computations into typed interaction contracts with executable golds. Selection replay recovers the released identities, artifact replay regenerates selected targets and intermediate outputs, and trajectory rescoring reproduces scores for stored executions.

XAI4HEAT provides partial portability evidence: shared protocol/runtime/scorer/alignment/replay operates on a second SCADA-like corpus with corpus-specific interaction/revision/reportability repairs. All 204 regenerated contracts pass canonical validation. The 41-row source audit records 994 comparisons, including metadata/link checks and recomputed numerical fields, with no mismatches.

\section{XAI4HEAT Porting Audit}
\label{sec:xai4heat-portability}

XAI4HEAT differs from BTS at the source boundary. BTS combines per-stream archive histories with graph-derived point, equipment, and location metadata, whereas XAI4HEAT provides row-oriented district-heating SCADA tables keyed by substation and channel. Its adapter maps substation--channel pairs into the normalized site/stream/point/equipment/timestamp/value representation. We regenerated the full XAI4HEAT benchmark under the final slot-consumption protocol contract, yielding 204 episodes with a 132/31/41 train/dev/test split across five temporal families.

The final audit identifies a compact cross-corpus adaptation layer. The normalized interface, candidate builders, dialogue renderer, clarification logic, tool runtime, protocol-dependency construction, deterministic scorer, trace normalizer/alignment, and replay logic are reused unchanged; task-family selection and evidence construction are configured. XAI4HEAT-specific interaction, revision, and reportability/quality semantics are concentrated in \path{scripts/canonical_repair_single_stream.py} and \path{scripts/canonical_repair_reportability.py}, which update task/interaction golds, dialogue state, verifier metadata, and tool calls where required by the corpus.

\begin{table*}[t]
\centering
\small
\setlength{\tabcolsep}{4pt}
\renewcommand{\arraystretch}{1.06}
\begin{tabularx}{0.97\textwidth}{>{\raggedright\arraybackslash}p{0.20\textwidth}>{\raggedright\arraybackslash}p{0.22\textwidth}>{\raggedright\arraybackslash}p{0.27\textwidth}X}
\toprule
Layer & BTS / shared role & XAI4HEAT result & Reuse status \\
\midrule
Source organization & Graph metadata + per-stream archives & Row-oriented SCADA tables + substation/channel mapping & Corpus-specific source adaptation \\
Normalized interface & Site, stream, point, equipment, time, value, lineage & Same normalized fields & Unchanged \\
Candidate builders & Shared telemetry task builders & Five supported temporal families & Unchanged; family selection configured \\
Interaction construction & Typed phases, revisions, reportability & Localized single-stream/reportability adaptation & Two corpus-specific modules \\
Protocol/runtime & Slot-consumption dependencies and read-only tools & 169/169 clarification edges slot-consistent; 134/134 independent-order controls accepted & Unchanged \\
Scoring/alignment/replay & Deterministic primary scorer and trace alignment & 204/204 canonical scorer pass; shared replay interface & Unchanged \\
Independent source check & Processed-source audit & 994 comparisons across 41 test rows: 952 exact + 42 tolerance matches, 0 mismatches & Validated within stated scope \\
\bottomrule
\end{tabularx}
\caption{Cross-corpus reuse on XAI4HEAT. Porting localizes corpus-specific interaction and reportability semantics in two adaptation modules while reusing the shared protocol, runtime, scoring, alignment, and replay stack.}
\label{tab:cross-corpus-portability}
\end{table*}

\begin{table}[t]
\centering
\small
\setlength{\tabcolsep}{4pt}
\renewcommand{\arraystretch}{1.05}
\begin{tabularx}{\linewidth}{>{\raggedright\arraybackslash}p{0.39\linewidth}X}
\toprule
Audit item & Final result \\
\midrule
Regenerated artifact & 204 rows; 132/31/41 train/dev/test \\
Row/split/family identity vs. archive & 204/204 unchanged \\
Provenance-only augmentation & 70 rows \\
Rows affected by XAI adaptation modules & 134 rows \\
Rows changing dialogue, phase, state, verifier, or tool-call fields & 44 rows \\
Clarification protocol edges & 169/169 slot-consistent \\
Independent clarification/point-order controls & 134/134 accepted \\
Final preflight / canonical scorer & 204/204 / 204/204 pass \\
Repeated deterministic outputs & 47/47 common row/static/compiler files byte-identical \\
Independent final-test source check & 994 comparisons on 41/41 rows; 952 exact + 42 tolerance matches; 0 mismatches \\

Porting result & Partial portability; corpus-specific downstream repairs remain \\
\bottomrule
\end{tabularx}
\caption{Final XAI4HEAT porting audit. The shared stack is reused with a localized corpus-specific interaction/reportability adaptation layer.}
\label{tab:xai4heat-portability}
\end{table}

\paragraph{Detailed final-build validation.}
The regenerated artifact contains 204 episodes with a 132/31/41 train/dev/test split under the final slot-consumption protocol semantics. All 204 contracts pass final preflight and canonical scoring. Across repeated builds, 47/47 common deterministic row/static/compiler outputs are byte-identical, and the logical tool-store contents are stable. Relative to the archived XAI4HEAT artifact, the localized adaptation layer updates pre-existing phase/gold fields on 134 rows and interaction/verifier/tool-call fields on 44 rows while preserving all 204 row IDs, splits, families, source linkages, and first source-task numeric targets.

The independent processed-source audit checks all 41 test rows: identities, timestamps/windows, observations/offsets, means, and quality coverage/gap metrics. Its 994 comparisons include metadata/link checks and numerical recomputation, not 994 numerical aggregations: 952 match exactly and 42 within stated tolerances, with no mismatches. Final policy commitments are not raw-source numerical facts.

The localized adaptation layer supplies corpus-specific temporal and reportability policies to shared scoring and protocol machinery. 

\paragraph{Artifact relationship.}
All 204 row identifiers, splits, families, source linkages, and first source-task numeric targets are preserved relative to the archived XAI4HEAT artifact. However, 134 rows differ in pre-existing phase/gold fields, and 44 rows also differ in dialogue, phase sequence, state/revision targets, verifier metadata, or tool-call operations/arguments. The remaining 70 rows preserve prior fields while adding semantic provenance metadata. These changes make the regenerated artifact a new interaction-contract revision. 

\paragraph{Final protocol validation.}
The shared slot-consumption rule generates 169 clarification dependencies, all of which are consistent with the slot consumed by the earliest dependent operation. No clarification edge is attached to independent point resolution, and all 134 independent clarification/point-order controls are accepted. The final contract preflight and canonical scorer both pass 204/204 rows.

\paragraph{Independent source check.}
For every test row, numerical recomputation reads original processed XAI4HEAT CSVs without stored gold as calculation input. The count above also includes identity/linkage comparisons. Aggregate and quality-ratio tolerance is 0.00005 (four-decimal rounding), and offset tolerance is 0.001 seconds. Pairwise comparison and ranking are unsupported families and outside this check.

\paragraph{Remaining corpus-specific semantics.}
Corpus-specific construction checks and regression tests validate the resulting temporal and reportability targets.

\end{document}